\documentclass[lettersize,journal]{IEEEtran}
\usepackage{amsmath,amsfonts}
\usepackage{algorithmic}
\usepackage{array}
\usepackage[caption=false,font=normalsize,labelfont=sf,textfont=sf]{subfig}
\usepackage{textcomp}
\usepackage{stfloats}
\usepackage{url}
\usepackage{cite}
\usepackage{verbatim}
\usepackage{graphicx}
\usepackage{listings}
\usepackage{xcolor}
\usepackage{graphicx}
\usepackage{svg}
\usepackage{multirow}
\usepackage{tabularx}
\lstdefinestyle{jsonstyle}{
  basicstyle=\ttfamily\scriptsize,
  frame=single,
  breaklines=true,
  columns=fullflexible,
  showstringspaces=false,
  xleftmargin=1mm,
  xrightmargin=1mm
}

\def\BibTeX{{\rm B\kern-.05em{\sc i\kern-.025em b}\kern-.08em
    T\kern-.1667em\lower.7ex\hbox{E}\kern-.125emX}}
\usepackage{balance}
\begin{document}
\title{COT-TTS: Audio Context-Aware Text-to-Speech with Chain-of-Thought Reasoning}

\author{
Weizhen~Bian$^{1}$,
Sitong~Cheng$^{1}$,
Rongxiu~Zhong$^{3,4}$,
Jiahao~Pan$^{1}$,
Liumeng~Xue$^{2}$,
Boyi~Kang$^{1}$,
Shilei~Zhang$^{3,4}$,
Jinglei~Liu$^{5}$,
Yue~Wang$^{5}$,
Junlan~Feng$^{3,\dagger}$,
Bei~Liu$^{1,\dagger}$,
and Wei~Xue$^{1,\dagger}$\\[0.4em]
$^{1}$The Hong Kong University of Science and Technology, Hong Kong SAR, China\\
$^{2}$Nanjing University\\
$^{3}$JIUTIAN Research, China Mobile, Beijing, China\\
$^{4}$The State Key Laboratory of Multimedia Information Processing, Peking University, Beijing, China\\
$^{5}$China Mobile (Hong Kong) Innovation Research Institute, Hong Kong SAR, China
\thanks{$^{\dagger}$Corresponding authors. 
fengjunlanit@chinamobile.com; 
beiliu@ust.hk; 
weixue@ust.hk.}
}

\markboth{}
{COT-TTS: Audio Context-Aware Text-to-Speech with Chain-of-Thought Reasoning}

\maketitle

\begin{abstract}
Recently, text-to-speech systems have made significant progress in speech expressiveness and controllability. However, the speaking style of generated speech typically relies on clear user-specified instructions. In natural conversations, speaking style should be naturally inferred from the preceding conversational context. Therefore, we propose COT-TTS, a context-aware, reasoning-based text-to-speech task. Given historical conversation audio, target text, and a reference speech, the system should comprehend the conversational context, infer an explicit intermediate reasoning, and finally synthesize the target speech with the specified timbre. To support this task, we constructed a large-scale bilingual conversational speech dataset comprising 9 million training samples, including a high-quality subset of 1 million samples. We further constructed a source-disjoint benchmark with 800 human-verified samples and established strong task-specific baselines. Additionally, we developed end-to-end autoregressive models with parameter sizes of 0.6B and 1.7B, generating emotion-labeled transcripts, editable speech style inferences, and speech tokens. Experimental results show that the proposed model achieves performance comparable to large-scale baseline systems with significantly fewer parameters. At the same time, the model performs well in terms of duration consistency and emotional consistency, and can generate appropriate emotional, stress, and rhythmic variations based on the conversational context. To facilitate future research, we will publicly release the data construction pipeline, dataset, trained models, and related resources. The demo page and additional resources are available at \url{https://luckybian.github.io/COT-TTS}.
\end{abstract}


\section{Introduction}

\IEEEPARstart{R}{ecent} advances in large language models and neural speech synthesis have made text-to-speech (TTS) systems increasingly controllable~\cite{xie2025towards}. Natural-language prompts are now widely used to control expressive speaking styles~\cite{yang2024instructtts}, and recent models provide finer control over speaker timbre, emotion, and speaking style~\cite{liao2026fish,zhou2025voxcpm,zhou2026indextts2}. Long-form conversational TTS systems have also been developed for scenarios such as podcasts, virtual interviews, and dialogue-based audiobooks~\cite{song2026borderless,xie2025soulx}. However, these systems still largely rely on users to specify how each sentence should be spoken. In dialogue, film dubbing, and audiobook narration, such instructions should ideally be inferred from the preceding context rather than provided manually for every utterance. This motivates audio context-aware reasoning TTS, where the model uses dialogue history to infer the intended speaking manner and synthesize the target speech accordingly. This capability is particularly important in practical applications. For example, spoken responses in emotional support dialogue should reflect the user's emotional state and the ongoing interaction~\cite{liu2021towards,hu2025chain}, while speech in film dubbing and virtual character applications should match the scene and character state~\cite{cong2025emodubber,li2025dialogueagents}. However, building such a system remains challenging for several reasons.

A major challenge is the lack of data that matches this formulation. Existing conversational speech datasets generally follow several directions. DailyDialog, DailyTalk, and their later extensions focus mainly on daily conversations, while adding speech recordings, emotion annotations, or style prompts~\cite{li2017dailydialog,lee2023dailytalk,liu2024emotion,jeon2025prompt}. However, their scenarios are often relatively simple, with limited speaker diversity and contextual variation. Other datasets collect speech from richer narrative and situated dialogue settings, providing more diverse scenes and expressive conditions~\cite{liu2024generative,bian2024emospeech,cheng2025dnaspeech,park2024let,chen2026actormind}. These resources usually provide contextual labels or role information, but rarely explain how the context leads to a particular speaking manner. Recent audio reasoning datasets introduce CoT-style supervision, but mainly focus on reasoning about what to answer rather than how a given utterance should be spoken~\cite{xie2025audio}. Overall, these datasets capture only part of the information required by audio context-aware reasoning TTS. They rarely provide explicit multi-dimensional reasoning paired with naturally occurring target speech at scale. This gap makes it difficult to train and evaluate models for inferring how a target utterance should be spoken from audio context. To address this gap, we develop a reproducible data construction pipeline and construct bilingual training samples together with a source-disjoint test benchmark.

Existing systems address speaking-style control through explicit instructions, reference speech, or editable acoustic attributes~\cite{zhou2026indextts2,liao2026fish,zhou2025voxcpm,zhou2026voxcpm2,hu2026voicesculptor}. These systems provide flexible control over speaker timbre and expressive style, but the intended speaking manner is generally specified by the user rather than inferred from historical dialogue. Existing ASR, LLM, and controllable TTS modules can also be combined into cascaded systems that first extract information from historical speech, infer speaking-style instructions, and then synthesize the target utterance~\cite{yang2026duplexcascade,an2024funaudiollm}. However, existing work has not established a unified cascaded framework for the complete task setting considered here. Moreover, the cascaded conversion process may lose paralinguistic information in the original speech, including non-verbal vocalizations (NVVs), rhythm, and emotional intensity. The final output also depends on whether the downstream TTS model can accurately realize the inferred speaking manner.

Several recent studies explore context-aware conversational speech synthesis from different perspectives. JELLY and Chain-Talker focus on emotional understanding and empathetic speech synthesis~\cite{cha2025jelly,hu2025chain}. Their explicit reasoning mainly concerns emotional or empathetic states, rather than broader reasoning about the intended speaking manner. CapTalk uses dialogue context to predict a CoT sequence of expressive attributes~\cite{su2026captalk}. However, its CoT mainly consists of attribute-level predictions. It does not explicitly explain why these attributes are appropriate given the dialogue context. Harness TTS uses structured contextual information to plan expressive speech control~\cite{shen2026harness}. It does not directly reason over historical dialogue audio. Beyond speech, Qwen-Music introduces Melody-CoT to plan melody tokens before full music generation~\cite{xu2026qwen}. This intermediate planning focuses on melody, rather than explicit reasoning about speaking manner from dialogue context. Related audio-language models, such as CogAudio-LLM and OSUM-EChat, reason over speech for dialogue response generation~\cite{zhao2026beyond,geng2025osum}. Their goal is mainly to generate appropriate dialogue responses, rather than to speak fixed target text with independent timbre control. In contrast, COT-TTS jointly considers historical dialogue audio, fixed target text, and reference speech. It explicitly reasons about the intended speaking manner before speech synthesis. The main contributions are summarized as follows:

\begin{itemize}

\item We formulate \emph{COT-TTS}, an audio context-aware reasoning TTS task. Given historical multi-speaker dialogue audio, target text, and reference speech, the model explicitly understands the dialogue history, reasons about the intended speaking manner, and synthesizes the target speech.

\item We develop a reproducible pipeline for constructing audio context-aware reasoning TTS data from long-form dialogue recordings. Starting from approximately 100K hours of Chinese and English dialogue audio, we construct 9M training samples, including a 1M high-quality subset.

\item We develop 0.6B- and 1.7B-parameter CoT-guided autoregressive models that perform contextual reasoning and speech synthesis. Compared with cascaded systems composed of multiple large models, our models use substantially fewer parameters while achieving comparable overall performance.

\item We construct a source-disjoint bilingual benchmark from approximately 3M test candidates. After automatic filtering and human validation, the benchmark contains 800 samples. We further establish an evaluation protocol combining objective metrics, LLM-based assessment, and human listening tests.

\item We publicly release the data construction pipeline, training and evaluation data, trained autoregressive models, and the corresponding training and inference code to support reproducible research.

\end{itemize}

\section{Task Formulation for Context-Aware Reasoning TTS}
\label{sec:task_formulation}

We define \emph{COT-TTS} as an audio context-aware reasoning TTS task. Given a continuous historical dialogue audio segment $H^{a}$, a target text $y$, and a reference speech utterance $r^{a}$, the goal is to generate an emotion-tagged historical transcript $\hat{\mathcal{T}}$, a reasoning analysis $\hat{\mathcal{C}}$, and the target speech $\hat{x}^{a}$:
\begin{equation}
(\hat{\mathcal{T}}, \hat{\mathcal{C}}, \hat{x}^{a}) = f_{\theta}(H^{a}, y, r^{a}).
\label{eq:task}
\end{equation}
Here, $H^{a}$ provides the preceding dialogue context, $y$ specifies the linguistic content to be spoken, and $r^{a}$ specifies the target speaker timbre. The predicted transcript $\hat{\mathcal{T}}$ converts the historical audio context into emotion-aware textual evidence. Based on this evidence and the original audio context, the reasoning analysis $\hat{\mathcal{C}}$ explains how the target sentence should be spoken across multiple dimensions. In this work, $\hat{\mathcal{C}}$ consists of five reasoning dimensions, including language act, scene semantics, cognition and motivation, expected communicative outcome, and emotional trajectory. The model also predicts duration and emotional expression intensity as explicit speaking attributes, followed by a structured final summary. The generated speech $\hat{x}^{a}$ is expected to preserve the content of $y$, follow the speaker characteristics provided by $r^{a}$, and express a speaking manner consistent with both the dialogue context and $\hat{\mathcal{C}}$.

In our constructed data, each sample provides a reference historical transcript $\mathcal{T}$, a reference reasoning annotation $\mathcal{C}$, and a ground-truth target speech utterance $x^{a}$, where $x^{a}$ is the naturally occurring next utterance from the original dialogue source. Thus, the task is grounded in naturally occurring contextual expression rather than manually designed style prompts. Unlike spoken dialogue response generation, the model does not decide what to say, since the target text $y$ is already given. Unlike conventional voice cloning TTS, the reference speech $r^{a}$ mainly specifies the target speaker timbre, while the speaking manner is expected to be inferred from the historical dialogue context $H^{a}$.

\section{Scalable Construction of the Large-Scale Bilingual COT-TTS Dataset}

\begin{figure*}[t]
    \centering
    \includegraphics[width=\textwidth]{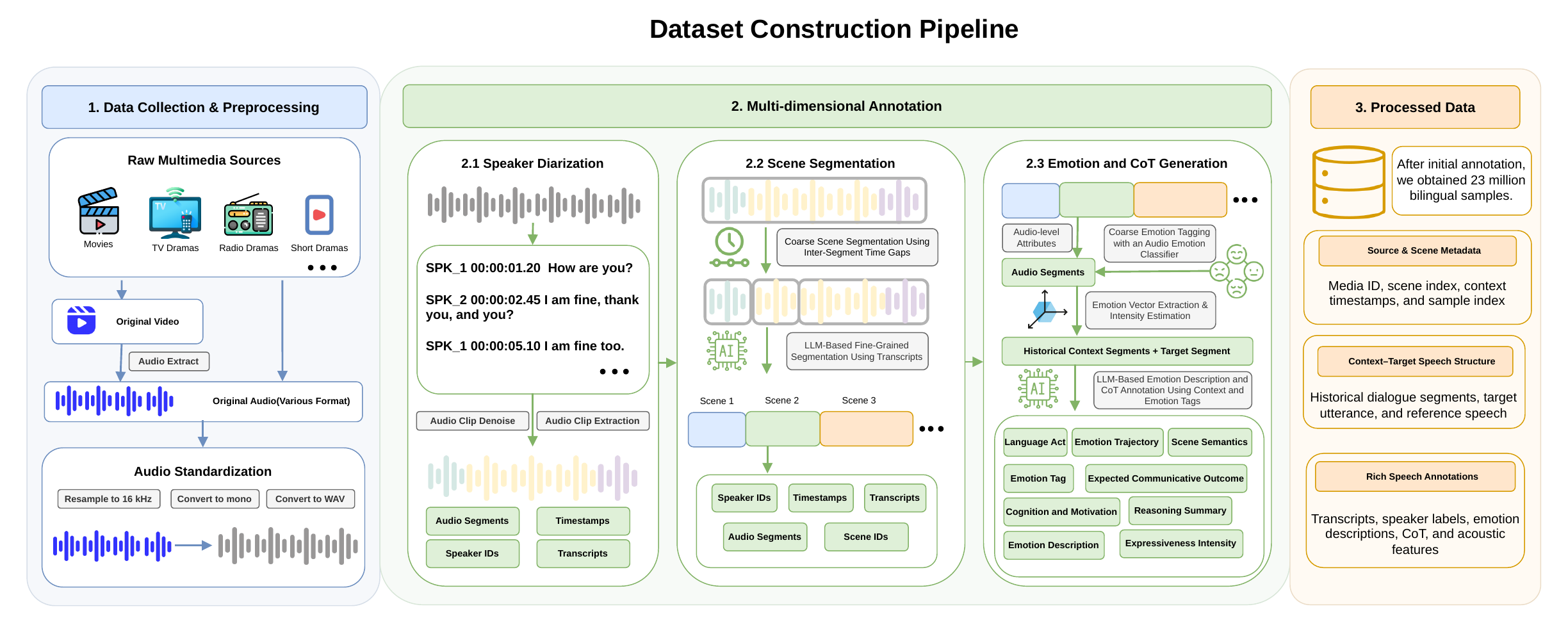}
    \caption{Overview of the COT-TTS data construction framework. The figure illustrates data collection and preprocessing, multi-dimensional annotation, and the resulting structured sample format for context-aware reasoning TTS.}
    \label{fig:data_pipeline}
\end{figure*}

Large-scale COT-TTS data must preserve dialogue context while providing explicit speaking-manner supervision and naturally occurring target speech. We therefore develop a scalable pipeline that converts Chinese and English dialogue recordings into task-aligned training samples. As illustrated in Fig.~\ref{fig:data_pipeline}, the pipeline includes data collection and preprocessing, multi-dimensional annotation, and structured sample construction. The following subsections describe the construction process and the resulting bilingual dataset.

\subsection{Data Collection and Preprocessing}

We collect Chinese and English dialogue-rich media from films, TV dramas, radio dramas, and short-form dramas. Compared with isolated read speech, these sources contain naturally occurring multi-speaker interactions, diverse scenes, and expressive speaking styles. The spoken lines are produced according to the surrounding plot, speaker relationships, and communicative intentions, making them suitable for studying how dialogue history influences the speaking manner of a target utterance. Detailed statements on data sources, copyright, licensing, and permitted usage will be released together with the dataset. For unified downstream processing, we extract the audio tracks from the collected media and normalize them into a consistent format. All recordings are converted to single-channel WAV files with a sampling rate of 16~kHz. These normalized long-form recordings are then used for subsequent speaker diarization, scene segmentation, and multi-dimensional annotation.

\subsection{Multi-dimensional Annotation}

After simple preprocessing, we first apply a pyannote-based processing pipeline to each long-form recording to obtain utterance-level timestamps, speaker labels, and aligned initial transcripts~\cite{Bredin23,Plaquet23}. Based on the predicted timestamps, we extract the corresponding utterance-level audio segments and apply FRCRN-based speech enhancement to reduce background noise~\cite{zhao2022frcrn}. 

We then annotate each recording with scene-level information. Scene segmentation is performed to provide a coherent local context for subsequent LLM-based annotation. This step is necessary because directly feeding a long episode into an LLM is inefficient and may introduce irrelevant contextual bias. We therefore first perform coarse segmentation using long pauses between utterances as boundaries. If a coarse segment still contains too many utterances, we further refine the scene boundaries using an LLM. The refinement is based on the transcript, speaker labels, and dialogue structure, producing more coherent dialogue scenes~\cite{deepseekai2025deepseekr1incentivizingreasoningcapability}. Each resulting segment is assigned a scene identifier. In this way, each target utterance is annotated mainly with respect to the historical dialogue within the same scene.

For emotion annotation, we avoid using only fixed emotion tags because a small set of categories may limit the expressiveness of the annotation. Directly using speech emotion captioning models can produce natural-language descriptions, but the generated captions may be inaccurate or unstable~\cite{xu2024secap}. Inspired by the coarse-to-fine emotion organization in OV-MER~\cite{lian2024ov}, we adopt a hierarchical annotation strategy. Each annotation is first assigned a coarse emotion category, followed by a finer-grained description. Specifically, we first use a speech emotion recognition model to determine the coarse emotion category of each target utterance~\cite{gao2023funasr}. We then project the target audio into an affective representation space containing arousal, dominance, and valence~\cite{wagner2023dawn}, and compute the emotional expression intensity as
\begin{equation}
I_{\mathrm{emo}} = 0.7 \cdot A + 0.3 \cdot D,
\label{eq:emo_intensity}
\end{equation}
where $A$ and $D$ denote arousal and dominance, respectively. The LLM subsequently combines the coarse emotion category, emotional expression intensity, and dialogue context to produce a concise emotion description. This design combines acoustic evidence with dialogue context. The SER model estimates the overall emotional direction from speech, while the LLM refines it using contextual cues, which helps with cases such as irony that are difficult to identify from text alone. We also extract several audio-level attributes for each target utterance to support subsequent data analysis and filtering. These attributes include audio duration, loudness, effective-speech ratio, audio quality score, and naturalness score. Together with the emotional expression intensity described above, these attributes provide quantitative measurements of the acoustic quality and expressive properties of each sample.

Finally, we annotate each target utterance with multi-dimensional CoT reasoning to explain its speaking manner from the historical context. Prior studies suggest that conversational expression depends on communicative function, context, and emotion development. Dialogue-act modeling characterizes the functional role of an utterance in an interaction~\cite{stolcke2000dialogue}. Contextual emotion studies further show that emotional expression is influenced by preceding events and speaker states~\cite{poria2019meld,ghosal2020cosmic}. Recent cognitive affective reasoning systems also organize speech-conditioned reasoning around affective perception, psychological inference, and response planning~\cite{zhao2026beyond}. Therefore, we generate the reasoning from the preceding three to five utterances within the same scene and organize it into five dimensions:

\begin{itemize}
\item \textbf{Language act analysis}: identifies the communicative function of the target utterance, such as expressing an opinion, narrating an event, or responding to another speaker.

\item \textbf{Scene semantic analysis}: describes the local situation in which the target utterance occurs and identifies the contextual information relevant to its expression.

\item \textbf{Cognition and motivation analysis}: infers the speaker's current cognitive state and the motivation behind the target utterance from the preceding dialogue.

\item \textbf{Expected communicative outcome analysis}: analyzes the communicative or emotional outcome that the speaker intends to produce through the target utterance.

\item \textbf{Emotional trajectory analysis}: explains how the speaker's emotion develops across the preceding dialogue and why the target utterance should carry a particular emotional tone.
\end{itemize}

We do not include speaker personality as an independent reasoning dimension. The speaker labels produced by pyannote may contain errors in scenes with overlapping speech or background noise, making it difficult to reliably aggregate utterances from the same character across different scenes. Moreover, personality traits are diverse and cannot be reliably inferred from a single local scene. We therefore focus on information that can be directly derived from the local dialogue context and associated with the target speech. In addition to the five reasoning dimensions, we include duration and emotional expression intensity as explicit speaking attributes. We then add a summary dimension that condenses the reasoning and speaking attributes into an overall speaking plan for the target utterance. The summary follows a structured pattern of ``because of [cause], in order to [goal], the speaker uses [emotion/style] to [action] and delivers [content].''

\subsection{Processed Data and Initial Analysis}

\begin{figure}[t]
    \centering
    \includegraphics[width=\columnwidth]{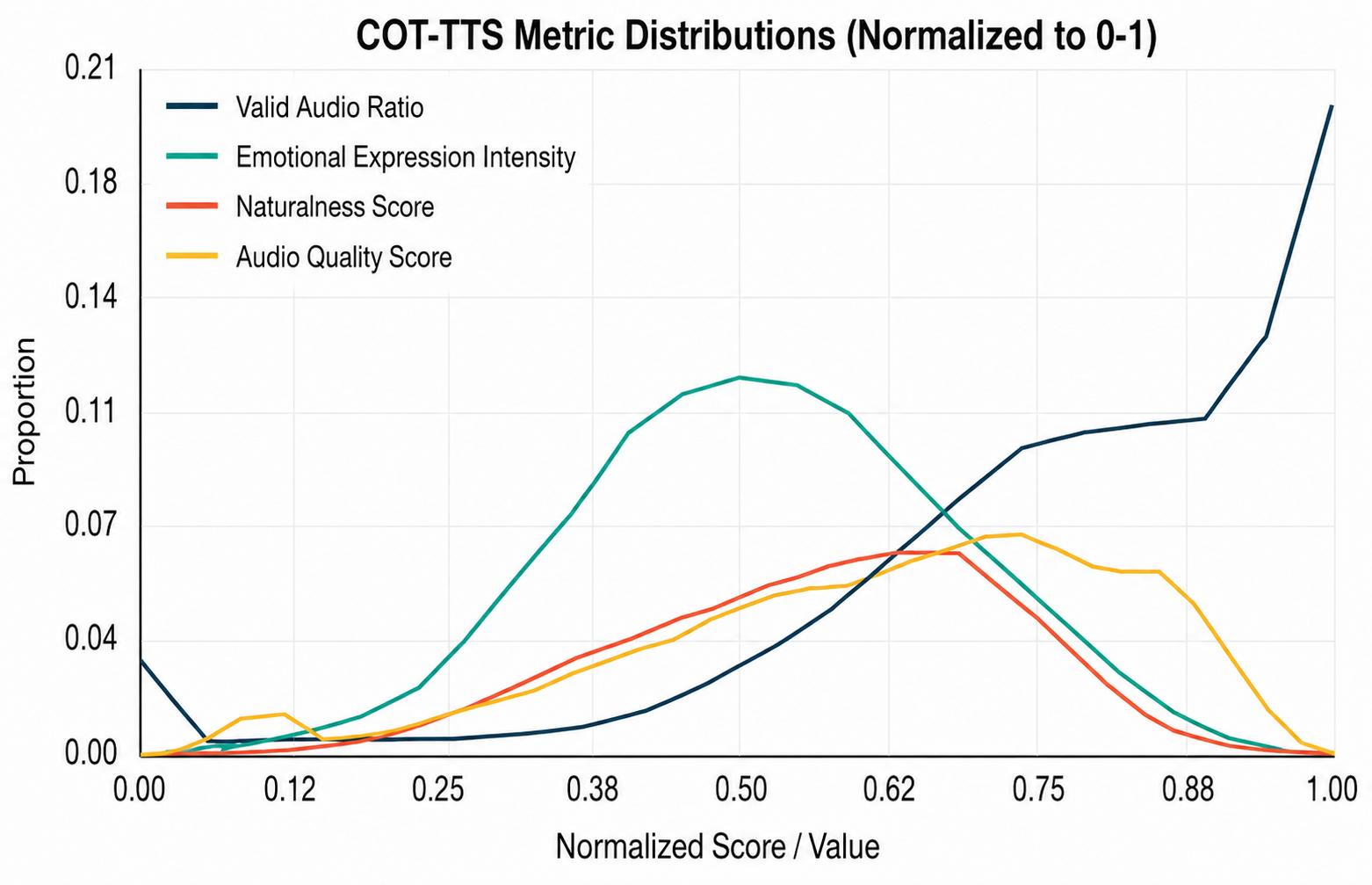}
    \caption{Distributions of the normalized audio quality score, naturalness score, target-audio effective-speech ratio, and emotional expression intensity over 50K randomly sampled examples.}
    \label{fig:data_analysis}
\end{figure}

Through the scalable construction pipeline, we process and annotate approximately 23M bilingual samples. Each sample contains historical multi-speaker dialogue audio, target speech, transcripts, and multi-dimensional annotations. To examine the quality of the automatically constructed data, we randomly sample 50K examples and evaluate their audio quality score, naturalness score, target-audio effective-speech ratio, and emotional expression intensity. For unified visualization, all metrics are normalized to the range of 0 to 1. The resulting distributions are shown in Fig.~\ref{fig:data_analysis}.

The distributions show that the automatically constructed data still vary considerably in quality. The effective-speech ratio is concentrated toward the high-value region, but a small portion still has very limited valid content. Emotional expression intensity is broadly distributed around the middle range, leaving many samples with relatively weak expression. The naturalness and audio quality scores also span a wide range, with a noticeable proportion of low-score samples. These variations can introduce unreliable supervision during training. We therefore apply a dedicated filtering and enhancement pipeline to remove low-quality samples and construct the final training and benchmark sets.

\subsection{Data Filtering and Enhancement}

\begin{figure}[t]
    \centering
    \includegraphics[width=\columnwidth]{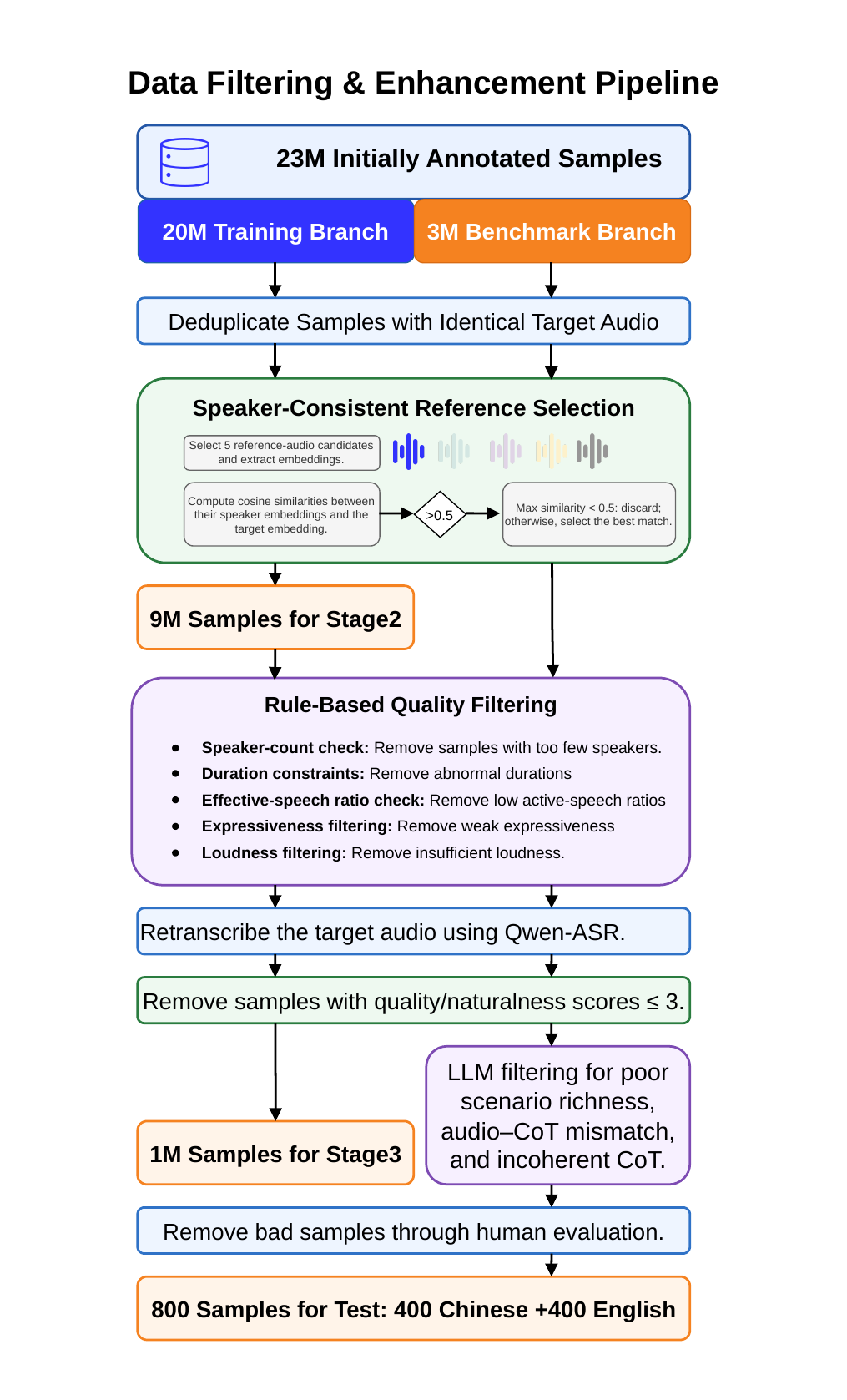}
    \caption{Overview of the COT-TTS data filtering and enhancement pipeline, including deduplication, reference selection, quality screening, ASR retranscription, LLM-based filtering, and human evaluation.}
    \label{fig:data_filtering}
\end{figure}

After the annotation stage shown in Fig.~\ref{fig:data_pipeline}, we split the approximately 23M initially annotated samples by source into a 20M training branch and a source-disjoint 3M benchmark branch. The two branches are subsequently processed independently to prevent source overlap between training and evaluation data. We first perform target-audio deduplication. During sample construction, the same target utterance may be paired with historical audio segments of different lengths. This results in multiple samples with identical target audio but different amounts of preceding dialogue context. We therefore identify groups of samples sharing the same target audio and randomly remove a portion of the redundant instances.

We then perform speaker-consistent reference selection for both branches. Speaker diarization is an efficient way to process large-scale conversational audio, but its outputs may contain speaker fragmentation, mixed-speaker segments, or incorrect speaker labels~\cite{ding2025kimi}. Since the historical audio $H^{a}$ is intended to contain multi-speaker dialogue, we focus speaker-consistency filtering on the target speech $x^{a}$ and reference speech $r^{a}$. For each target utterance, we retrieve five reference candidates with the same speaker and extract speaker embeddings~\cite{zeinali2019rvector,wang2023wespeaker}. We then compute the cosine similarity between the target speech embedding and each candidate embedding. If the maximum similarity is below 0.5, the sample is discarded; otherwise, the candidate with the highest similarity is selected as the final reference speech. This step removes samples with unreliable speaker assignments while ensuring that the retained reference speech matches the target speaker. After deduplication and reference selection, the training branch contains approximately 9M samples, which are used for Stage-2 training.

We further apply rule-based filtering according to the structural, acoustic, and expressive properties of each sample. Samples with too few speakers, abnormal audio durations, insufficient effective-speech ratios, weak emotional expression, or insufficient loudness are removed. We also discard samples whose audio quality or naturalness scores are no higher than 3~\cite{baba2024t05,cumlin2024dnsmos}. Because the initial transcripts produced during large-scale diarization may contain missing words or recognition errors, we re-transcribe the retained target audio using Qwen3-ASR~\cite{Qwen3-ASR}. After these filtering and enhancement steps, we obtain a high-quality subset of approximately 1M samples for Stage-3 training.

The benchmark branch is processed using the same filtering pipeline described above. To ensure the reliability of the benchmark, we further evaluate the remaining candidates using both LLM-based scoring and human assessment~\cite{Qwen3-Omni,deepseekai2025deepseekr1incentivizingreasoningcapability}. Each sample is rated on a five-point scale in terms of scenario richness, CoT logical accuracy, audio--CoT consistency, and transcript accuracy. We retain only samples with an average score above 4.3 and a score above 4.0 for every individual criterion. The statistics of the resulting data subsets are summarized in Table~\ref{tab:data_statistics}.

\begin{table}[t]
\centering
\caption{Statistics of the filtered COT-TTS data subsets.}
\label{tab:data_statistics}
\small
\begin{tabular*}{\columnwidth}{@{\extracolsep{\fill}}lcccc@{}}
\hline
\textbf{Subset} & \textbf{Language} & \textbf{Hours} &
\textbf{Sample Ratio} & \textbf{Samples} \\
\hline
\multirow{3}{*}{Stage-2}
& English & $\sim$21K & 51.73\% & $\sim$4.64M \\
& Chinese & $\sim$29K & 48.27\% & $\sim$4.33M \\
& Total   & $\sim$50K & 100\% & $\sim$9.0M \\
\hline
\multirow{3}{*}{Stage-3}
& English & $\sim$2.5K & 56\% & $\sim$0.56M \\
& Chinese & $\sim$3.0K & 44\% & $\sim$0.44M \\
& Total   & $\sim$5.5K & 100\% & $\sim$1.0M \\
\hline
\multirow{3}{*}{Test}
& English & $\sim$3.3 & 50\% & 400 \\
& Chinese & $\sim$3.3 & 50\% & 400 \\
& Total   & $\sim$6.7 & 100\% & 800 \\
\hline
\end{tabular*}
\end{table}

\section{End-to-End CoT-Guided Autoregressive Model}

\begin{figure*}[t]
    \centering
    \includegraphics[width=1.9\columnwidth]{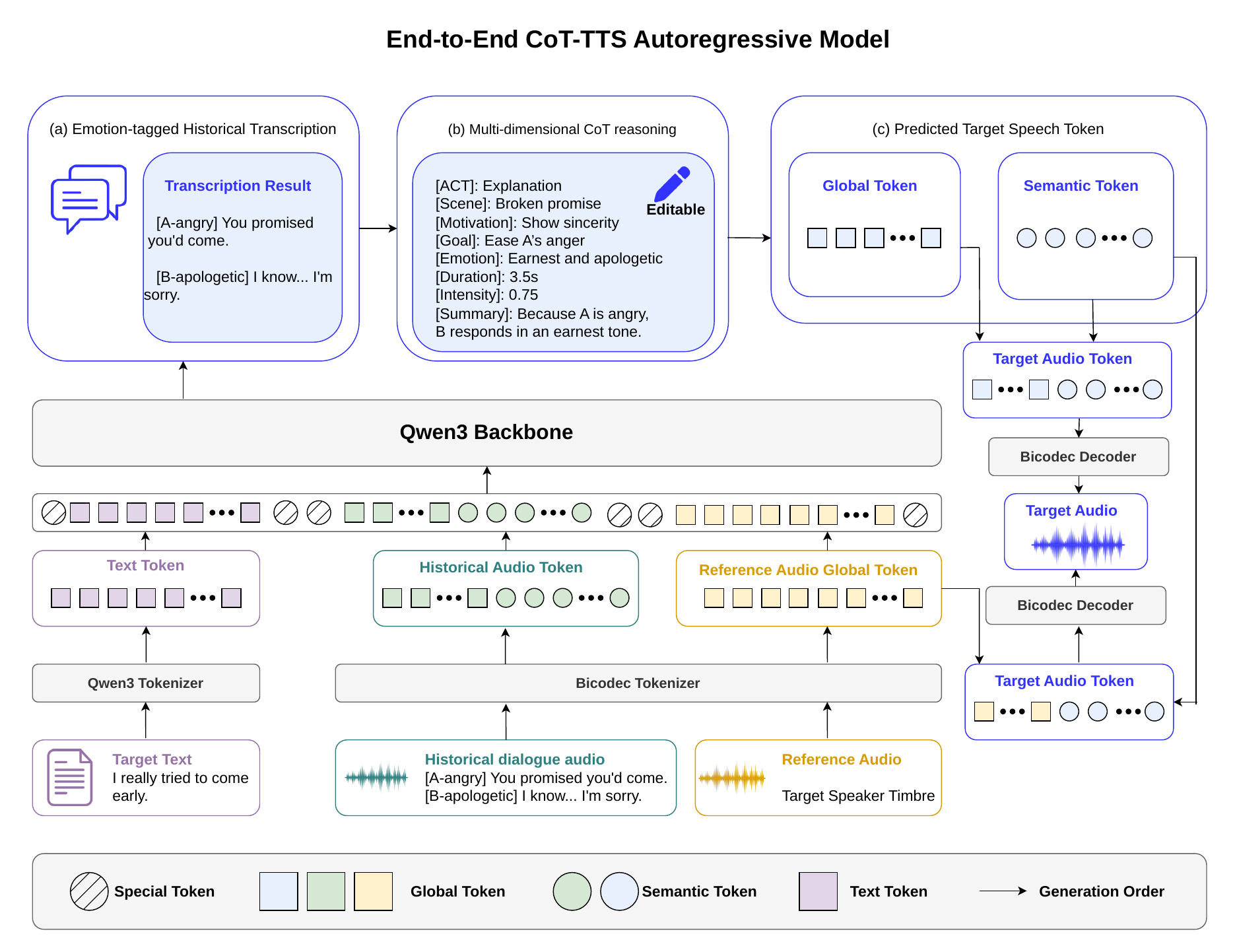}
    \caption{Overview of the end-to-end CoT-guided autoregressive model. Historical dialogue audio, target text, and reference speech are encoded into a unified token sequence. The model sequentially generates an emotion-tagged historical transcript, multi-dimensional CoT reasoning, and target speech tokens. The intermediate CoT can be inspected and edited before speech synthesis. For waveform reconstruction, the predicted semantic tokens are combined with either the reference or predicted global tokens.}
    \label{fig:e2e_model}
\end{figure*}

\subsection{Unified Sequence Formulation}

We implement the model using Qwen3 backbones with 0.6B and 1.7B parameters. Text and CoT reasoning are represented with the Qwen tokenizer, while speech is encoded into discrete tokens for unified autoregressive modeling. Continuous speech representations preserve fine-grained acoustic information but do not naturally fit the discrete next-token prediction paradigm of LLMs~\cite{yan2025ming}. Multi-codebook neural codecs provide high-quality reconstruction but require multiple parallel token streams, increasing the complexity of autoregressive generation~\cite{zeghidour2021soundstream,defossez2022high}. We therefore adopt BiCodec from Spark-TTS~\cite{wang2025spark}, which represents speech with a fixed-length global token block and a single time-varying semantic token sequence. This representation allows historical dialogue audio, reference speech, target text, CoT reasoning, and target speech to be organized into a unified autoregressive sequence. As illustrated in Fig.~\ref{fig:e2e_model}, the model follows the output order defined in Section~\ref{sec:task_formulation}, sequentially generating the historical transcript, CoT reasoning, and target speech tokens. The intermediate CoT can be inspected or edited before speech synthesis.

For a speech segment $u^{a}$, BiCodec produces 32 global tokens and a sequence of semantic tokens:
\begin{equation}
\mathrm{BiCodec}(u^{a})=(g_u,s_u),
\end{equation}
where $g_u \in \mathbb{Z}^{32}$ denotes the global tokens and $s_u$ denotes the semantic token sequence. The global tokens encode compact utterance-level information, including speaker characteristics, while the semantic tokens mainly represent linguistic content.

The historical dialogue audio $H^{a}$ consists of the preceding utterances in the same scene:
\begin{equation}
H^{a}={h_1^{a},h_2^{a},\ldots,h_N^{a}}.
\end{equation}
For each utterance, we extract BiCodec tokens as
$\mathrm{BiCodec}(h_i^{a})=(g_{h_i},s_{h_i})$. Retaining both global and semantic tokens for every utterance would introduce multiple speaker-dependent global token blocks and require reliable utterance segmentation during inference. Since the model is intended to directly process a continuous historical audio segment, we instead construct its representation as
\begin{equation}
H^{a}\Rightarrow
[g_{h_1},s_{h_1},s_{h_2},\ldots,s_{h_N}].
\end{equation}
Only one global token block is retained, while the semantic sequences of all historical utterances are concatenated to preserve the dialogue content. In our preliminary experiments, removing the global tokens entirely degraded the model's transcription and understanding of historical audio. We therefore retain one global token block as an acoustic anchor to support semantic understanding of the historical audio. This design avoids introducing a separate speaker-dependent global token block for every historical utterance. It also makes the training representation more similar to the representation obtained by encoding continuous historical audio during inference.

The reference speech $r^{a}$ is used to provide the target-speaker timbre. We extract its BiCodec representation as $\mathrm{BiCodec}(r^{a})=(g_r,s_r)$ but retain only the 32 global tokens:
\begin{equation}
r^{a}\Rightarrow g_r,\qquad g_r\in\mathbb{Z}^{32}.
\end{equation}
For the target speech $x^{a}$, the model autoregressively predicts both global and semantic tokens:
\begin{equation}
x^{a}\Rightarrow(\hat{g}_x,\hat{s}_x).
\end{equation}
The target waveform can then be reconstructed in two ways. The first directly decodes the complete token sequence predicted by the model:
\begin{equation}
\hat{x}^{a}_{\mathrm{pred}}
=
\mathrm{BiCodecDecoder}(\hat{g}_x,\hat{s}_x).
\end{equation}
The second combines the global tokens extracted from the reference speech with the predicted target semantic tokens:
\begin{equation}
\hat{x}^{a}_{\mathrm{ref}}
=
\mathrm{BiCodecDecoder}(g_r,\hat{s}_x).
\end{equation}
The first mode decodes the complete target representation predicted by the model, while the second mode directly uses the reference global tokens for speaker conditioning. We adopt the second mode as the default reconstruction strategy.

\subsection{Three-stage Training Strategy}

\begin{figure*}[t]
    \centering
    \includegraphics[width=1.8\columnwidth]{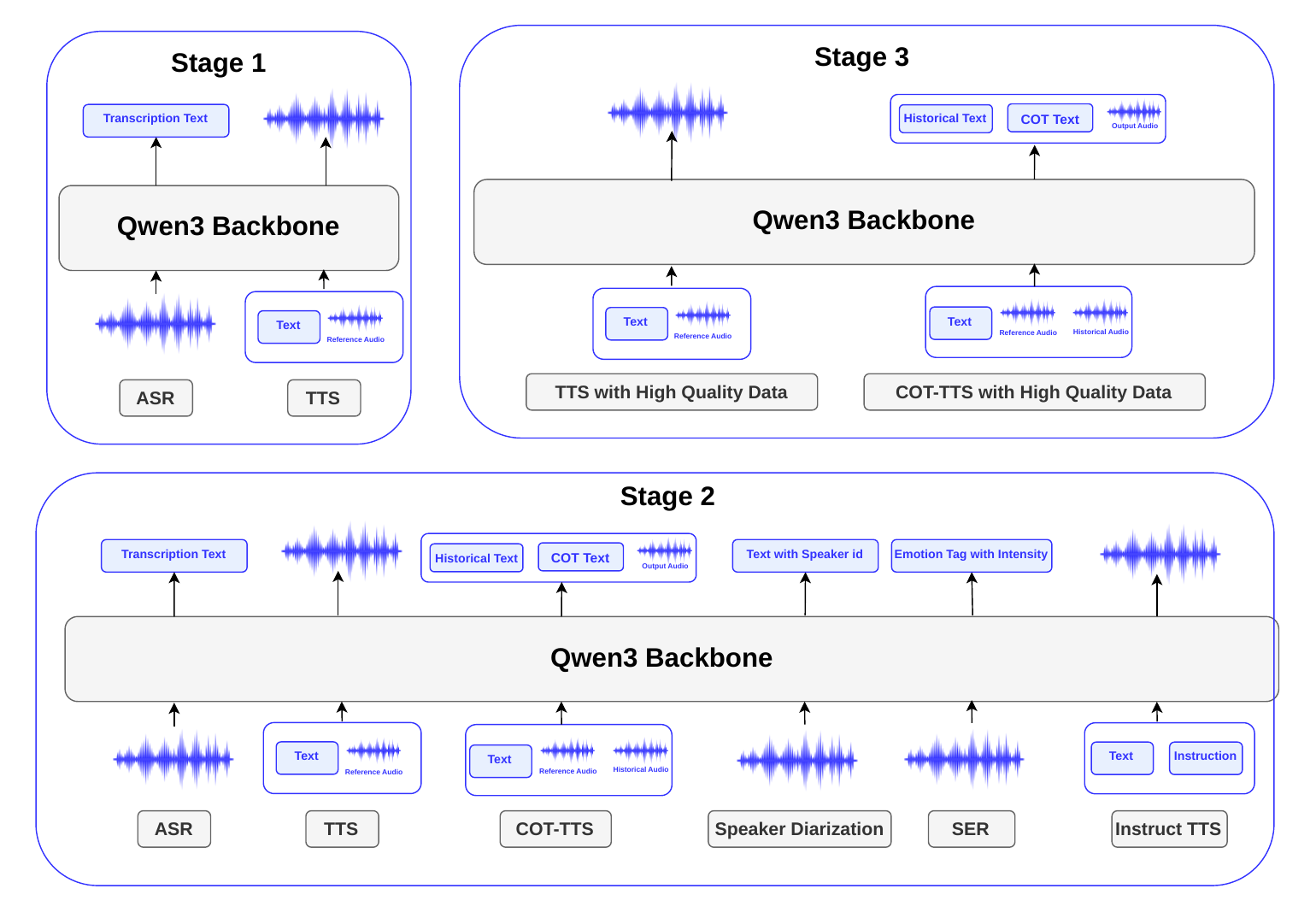}
    \caption{Overview of the three-stage training strategy for our end-to-end autoregressive models for COT-TTS. The figure summarizes the training data, model inputs, and outputs at each stage.}
    \label{fig:training_strategy}
\end{figure*}

We train the model in three stages using the VeOmni framework with full-parameter fine-tuning. As illustrated in Fig.~\ref{fig:training_strategy}, the first stage establishes text--speech modality alignment, the second stage introduces the main COT-TTS task through multi-task training, and the third stage performs quality-oriented refinement using high-quality TTS and COT-TTS data. All tasks are converted into a unified sequence format with predefined input and output fields separated by task-specific special tokens. These special tokens are newly added to the tokenizer to identify different functional regions. For each task, the input tokens are used as context, while the next-token prediction loss is computed only on the target output tokens. Detailed training hyperparameters and configurations are provided in the released code.

The first stage focuses on modality alignment. We use ASR and TTS tasks to equip the Qwen3 backbone with basic speech understanding and synthesis abilities. For ASR, the model takes speech tokens as input and predicts the corresponding text tokens. For TTS, the model takes text tokens and reference global tokens as input and predicts the target speech tokens. Following the Spark-TTS setting, we use approximately 140K hours of filtered open-source speech data to construct the ASR and TTS tasks~\cite{wang2025spark}. This stage establishes the basic alignment between Qwen text tokens and BiCodec speech tokens.

The second stage introduces the main COT-TTS task using the 9M Stage-2 training samples. Given the historical dialogue audio $H^{a}$, target text $y$, and reference global tokens $g_r$, the model predicts the emotion-tagged historical transcript $\mathcal{T}$, CoT speaking-manner reasoning $\mathcal{C}$, and target speech tokens $(g_x,s_x)$. Since the main task jointly requires speech understanding, contextual reasoning, timbre conditioning, and speech synthesis, we further introduce several auxiliary tasks to strengthen the corresponding abilities. As shown in Fig.~\ref{fig:training_strategy}, the auxiliary tasks include ASR, conventional TTS, speaker diarization, speech emotion recognition, and instruction-style TTS. The speaker diarization task takes dialogue audio as input and predicts transcripts with speaker identifiers. The speech emotion recognition task predicts the emotion category and expression intensity from speech. The instruction-style TTS task takes target text and an instruction as input and predicts the corresponding speech tokens. During this stage, the main COT-TTS task and auxiliary tasks are mixed at a ratio of 7:3, with the auxiliary tasks sampled uniformly. To improve robustness, we randomly drop 0--10\% of the input tokens with a probability of 50\%. This stage establishes the model's initial ability to perform audio context-aware and speech synthesis.

The third stage refines the model using 1M high-quality COT-TTS samples and 1M high-quality TTS samples. The high-quality TTS data improve the acoustic quality and stability of the generated speech, while the COT-TTS data further adapt the model to the target task distribution. All auxiliary tasks used in the second stage are removed. By jointly training on these two high-quality subsets, this stage improves output speech quality while shifting the overall model distribution toward COT-TTS. The training configurations are summarized in Table~\ref{tab:training_configuration}.

\begin{table}[!htbp]
\centering
\caption{Training configurations of the three stages. Epochs are listed for the 0.6B/1.7B models.}
\label{tab:training_configuration}
\scriptsize
\setlength{\tabcolsep}{2pt}
\begin{tabularx}{\columnwidth}{cXccc}
\hline
\textbf{Stage} & \textbf{Data} & \textbf{Ratio} & \textbf{Zh:En} & \textbf{Epochs} \\
\hline
1 & ASR (70K h), TTS (70K h) & 1:1 & 1:1 & 1|1 \\
2 & COT-TTS (9M) + Auxiliary tasks & 7:3 & 48:52 & 2|3 \\
3 & COT-TTS (1M) + TTS (1M) & 1:1 & 44:56 & 1|1 \\
\hline
\end{tabularx}
\end{table}

\subsection{Inference with Editable CoT}

During inference, the model follows the same autoregressive order as in training. As illustrated in Fig.~\ref{fig:e2e_model}, the model first generates the emotion-tagged historical transcript $\hat{\mathcal{T}}$ and speaking-manner reasoning $\hat{\mathcal{C}}$. Since the reasoning is generated before speech synthesis, decoding can be paused after the CoT stage to inspect or edit the intermediate reasoning. Formally, the original reasoning can be replaced with an edited version $\hat{\mathcal{C}}'$:
\begin{equation}
\hat{\mathcal{C}} \rightarrow \hat{\mathcal{C}}',
\end{equation}
after which autoregressive decoding continues as
\begin{equation}
(\hat{g}_x,\hat{s}_x)
\sim
p_{\theta}
\left(
\cdot
\mid
H^{a}, y, g_r, \hat{\mathcal{T}}, \hat{\mathcal{C}}'
\right).
\end{equation}
In practice, the editable mode is intended for moderate revisions of the generated CoT while preserving its overall reasoning structure. Large modifications may disrupt the learned dependency between the reasoning and subsequent speech tokens, leading to reduced speech quality and naturalness. As described above, although both $\hat{g}_x$ and $\hat{s}_x$ are autoregressively predicted, the reference global tokens $g_r$ are combined with $\hat{s}_x$ for final waveform reconstruction to provide reference-speaker conditioning.

\section{Experimental Setup}

\subsection{Evaluation Protocol}

All systems are evaluated on the source-disjoint COT-TTS benchmark described above, which contains 800 test samples, including 400 Chinese and 400 English samples. Evaluation of conversational and expressive speech synthesis commonly combines automatic metrics with subjective listening tests. Existing conversational TTS studies mainly examine whether dialogue context improves speech naturalness and prosody~\cite{xue2023m}, while instruction-based benchmarks evaluate whether generated speech follows explicit style descriptions~\cite{huang2025instructttseval}. More recent studies further consider the plausibility and controllability of intermediate speaking-style reasoning~\cite{su2026captalk}. Since COT-TTS requires both context-dependent reasoning and expressive speech synthesis, we evaluate each system from three complementary perspectives: objective speech quality, LLM-based reasoning quality, and human subjective assessment.

The objective metrics evaluate speech quality, intelligibility, and acoustic consistency. We use UTMOSv2 to estimate speech naturalness~\cite{baba2024t05}, and DNSMOSPro to estimate overall perceptual speech quality~\cite{cumlin2024dnsmos}. Since the target text $y$ is fixed and provided to all systems, we use character error rate (CER) for Chinese samples and word error rate (WER) for English samples to evaluate whether the synthesized speech preserves the intended linguistic content. We further measure emotional-expression intensity, duration, and F0 consistency between the synthesized speech $\hat{x}^{a}$ and the ground-truth target speech $x^{a}$. For emotional-expression consistency, we use the VAD-based emotional expression intensity $I(\cdot)$ defined in Eq.~\eqref{eq:emo_intensity} and compute
\begin{equation}
S_{\mathrm{emo}}
=
\max\left(
0,\,
1-\frac{\left|I(\hat{x}^{a})-I(x^{a})\right|}{R_I}
\right),
\end{equation}
where $R_I=I_{\max}-I_{\min}$ denotes the valid range of the emotional expression intensity score. A higher $S_{\mathrm{emo}}$ indicates closer agreement with the expressive intensity of the ground-truth speech. Duration consistency is measured by the absolute duration error
\begin{equation}
E_{\mathrm{dur}}
=
\left|
D(\hat{x}^{a})-D(x^{a})
\right|,
\end{equation}
where $D(\cdot)$ denotes speech duration in seconds, and a lower $E_{\mathrm{dur}}$ indicates better temporal consistency. For F0 consistency, we extract the F0 contours of $\hat{x}^{a}$ and $x^{a}$, resample them to the same length, and retain only frames with valid F0 values in both sequences. We then compute the Pearson correlation coefficient
\begin{equation}
S_{\mathrm{F0}}
=
\frac{
\sum_{t=1}^{T}
(\hat{f}_t-\bar{\hat{f}})
(f_t-\bar{f})
}{
\sqrt{\sum_{t=1}^{T}(\hat{f}_t-\bar{\hat{f}})^2}
\sqrt{\sum_{t=1}^{T}(f_t-\bar{f})^2}
},
\end{equation}
where $\hat{f}_t$ and $f_t$ denote the valid F0 values of $\hat{x}^{a}$ and $x^{a}$ after resampling, respectively, and $\bar{\hat{f}}$ and $\bar{f}$ denote their corresponding mean values. A higher $S_{\mathrm{F0}}$ indicates stronger consistency in pitch variation.

\begin{figure*}[t]
    \centering
    \includegraphics[width=1.9\columnwidth]{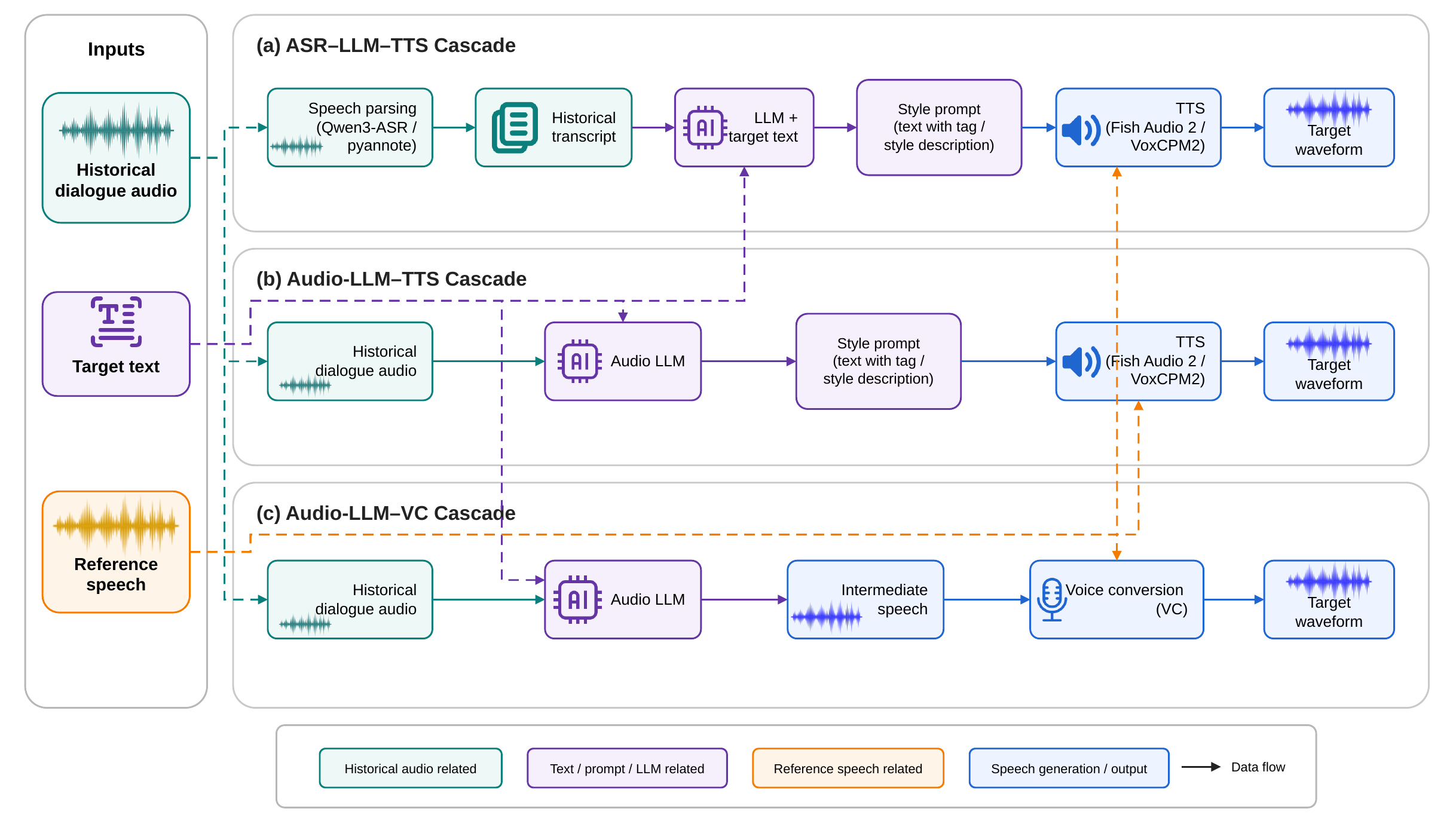}
    \caption{Overview of the three cascaded baseline architectures. (a) The three-stage ASR/diarization--LLM--TTS pipeline first converts historical dialogue audio into text, then infers the speaking manner with an LLM, and finally synthesizes the target speech using a controllable TTS model. (b) The two-stage AudioLLM--TTS pipeline directly analyzes the historical audio with an AudioLLM and uses the inferred speaking manner to control downstream TTS. (c) The two-stage AudioLLM--VC pipeline directly generates an intermediate target speech with an AudioLLM and subsequently applies voice conversion to condition the output on the reference speaker.}
    \label{fig:baseline_systems}
\end{figure*}

For reasoning evaluation, we employ an LLM judge to provide three component scores~\cite{deepseekai2025deepseekr1incentivizingreasoningcapability}: historical understanding, CoT internal logic, and information richness, each ranging from 0 to 5. Historical understanding evaluates whether the generated reasoning correctly understands the information contained in the historical dialogue. CoT internal logic evaluates whether the generated CoT is internally logical and consistent. Information richness evaluates the diversity and amount of information in the CoT that can be meaningfully used to guide subsequent speech synthesis. To improve evaluation accuracy, we follow reasoning-based LLM evaluation frameworks~\cite{wang2026tts,cheng2026steb} and require the judge to first provide a scoring rationale before assigning the component scores, reducing the risk of producing an arbitrary score followed by a post-hoc justification. To further improve stability and reliability, each sample is independently evaluated three times. For each component, the three independently obtained scores are aggregated using a consensus-based strategy. If all three scores are identical, the common score is retained. If two scores agree, they are treated as the consensus; an outlier differing from the consensus by at least 2 points is discarded, whereas smaller disagreements are resolved by averaging the three scores. If all three scores are different, their median is used as the final component score. The rationale associated with the retained or consensus score is preserved for subsequent analysis. This strategy reduces the influence of stochastic variation and occasional anomalous judgments from a single LLM evaluation. Based on the three aggregated component scores, we further compute a composite score to summarize the overall reasoning quality. To prevent logically correct but low-information reasoning from receiving a high composite score, information richness is additionally used as a weighting factor:
\begin{equation}
S_{\mathrm{comp}}
=
\frac{S_{\mathrm{info}}}{5}
\cdot
\frac{
S_{\mathrm{hist}}+S_{\mathrm{logic}}+S_{\mathrm{info}}
}{3},
\end{equation}
where $S_{\mathrm{hist}}$, $S_{\mathrm{logic}}$, and $S_{\mathrm{info}}$ denote the aggregated historical-understanding, CoT-internal-logic, and information-richness scores, respectively. This formulation penalizes reasoning that is superficially correct or logically coherent but provides little useful information for determining how the target utterance should be spoken. For example, a generic explanation such as ``the speaker needs to respond because a question was asked'' may be logically valid but contains little information useful for speech synthesis.

In addition to the automatic metrics and LLM-based evaluation, we conduct a blind human subjective evaluation. Raters are presented with the historical dialogue audio, the generated CoT, and the corresponding synthesized speech, and assign an overall score from 1 to 5. The evaluation considers the appropriateness of the speaking manner with respect to the historical dialogue context, the consistency between the speech and CoT reasoning, and the overall perceptual quality. The final Human MOS is obtained by averaging the scores across all evaluated samples and raters.

\subsection{Compared Systems}

The COT-TTS task requires historical dialogue understanding, speaking-manner reasoning, target speech synthesis, and reference-speaker conditioning. Since existing off-the-shelf systems typically support only a subset of these capabilities, we construct three types of cascaded baselines, as illustrated in Fig.~\ref{fig:baseline_systems}.

The first group follows an \textbf{ASR--LLM--TTS} pipeline. Historical dialogue audio is transcribed using either Qwen3-ASR~\cite{Qwen3-ASR} or diarization-based speech parsing~\cite{Plaquet23,Bredin23}, after which Qwen3-30B-A3B analyzes the dialogue context and generates speaking-manner instructions~\cite{qwen3technicalreport}. Fish Audio 2~\cite{liao2026fish} and VoxCPM2~\cite{zhou2026voxcpm2} are used as controllable TTS backends, resulting in four baselines: three-qwen\_asr-a3b-fish, three-qwen\_asr-a3b-voxcpm, three-dia-a3b-fish, and three-dia-a3b-voxcpm.

The second group follows an \textbf{AudioLLM--TTS} pipeline, where Qwen3-Omni-30B-A3B directly processes the historical dialogue audio and generates the speaking-manner instruction~\cite{Qwen3-Omni}. Combined with Fish Audio 2 and VoxCPM2 as downstream TTS models, this yields two-a3b-fish and two-a3b-voxcpm, respectively. The third baseline follows an \textbf{AudioLLM--VC} pipeline, where Qwen3-Omni-30B-A3B generates an intermediate target speech and Seed-VC converts it toward the reference-speaker timbre~\cite{Qwen3-Omni,liu2024zero}. This system is denoted as two-qwen3omni-seedvc. All these systems are compared with our 0.6B and 1.7B end-to-end models trained with Qwen3 backbones~\cite{qwen3technicalreport}.

\begin{table*}[t]
\centering
\caption{Main results on the English COT-TTS benchmark. ``$\uparrow$'' indicates that higher is better, and ``$\downarrow$'' indicates that lower is better.}
\label{tab:main_results_en}
\resizebox{\textwidth}{!}{
\begin{tabular}{
l c
@{\hspace{8pt}}
c c c c c c
@{\hspace{12pt}}
c c c c
@{\hspace{12pt}}
c
}
\hline
\textbf{System}
& \textbf{Model Size}
& \multicolumn{6}{c}{\textbf{Speech Quality and Acoustic Consistency}}
& \multicolumn{4}{c}{\textbf{Context-Aware Reasoning Quality}}
& \multicolumn{1}{c}{\textbf{Human Evaluation}} \\
\cline{3-8}
\cline{9-12}
\cline{13-13}
&
& \shortstack{\textbf{UTMOSv2}\\$\uparrow$}
& \shortstack{\textbf{DNSMOSPro}\\$\uparrow$}
& \shortstack{\textbf{WER}\\$\downarrow$}
& \shortstack{\textbf{Emo. Cons.}\\$\uparrow$}
& \shortstack{\textbf{Dur. Err.}\\$\downarrow$}
& \shortstack{\textbf{F0 Corr.}\\$\uparrow$}
& \shortstack{\textbf{Hist. Und.}\\$\uparrow$}
& \shortstack{\textbf{CoT Logic}\\$\uparrow$}
& \shortstack{\textbf{Eff. Info. Richness}\\$\uparrow$}
& \shortstack{\textbf{LLM Composite}\\$\uparrow$}
& \shortstack{\textbf{Human MOS}\\$\uparrow$} \\
\hline

three-dia-a3b-fish
& 36.2B
& 3.109
& 3.287
& \textbf{0.012}
& 0.910
& 5.866
& \textbf{0.227}
& \textbf{4.075}
& 4.587
& 3.574
& 2.915
& 3.050 \\

three-dia-a3b-voxcpm
& 34.2B
& 2.714
& 3.240
& 1.307
& 0.952
& 12.495
& 0.117
& 4.053
& 4.601
& \textbf{4.202}
& \textbf{3.601}
& 3.450 \\

three-qwen\_asr-a3b-fish
& 36.2B
& 3.102
& 3.275
& 0.092
& 0.917
& 5.267
& 0.217
& 3.816
& 4.759
& 2.941
& 2.258
& 3.500 \\

three-qwen\_asr-a3b-voxcpm
& 34.2B
& 2.663
& 3.223
& 0.165
& 0.944
& 5.852
& 0.133
& 3.820
& 4.753
& 4.141
& 3.510
& 3.300 \\

two-a3b-fish
& 39.0B
& \textbf{3.138}
& 3.269
& 0.090
& 0.920
& 5.112
& 0.213
& 3.689
& \textbf{4.836}
& 2.644
& 1.969
& 3.050 \\

two-a3b-voxcpm
& 37.0B
& 2.638
& 3.194
& 0.110
& 0.945
& 5.388
& 0.124
& 3.667
& 4.812
& 3.975
& 3.300
& 3.400 \\

two-qwen3omni-seedvc
& $\sim$35.7B
& 3.007
& \textbf{4.240}
& 0.101
& 0.953
& 5.535
& 0.146
& --
& --
& --
& --
& 3.250 \\

\hline

COT-TTS-0.6B
& 0.6B
& 2.943
& 3.180
& 0.041
& \textbf{0.954}
& 1.155
& 0.168
& 3.445
& 4.558
& 3.110
& 2.304
& \textbf{4.150} \\

COT-TTS-1.7B
& 1.7B
& 2.960
& 3.207
& 0.054
& \textbf{0.954}
& 1.063
& 0.212
& 3.573
& 4.720
& 3.070
& 2.326
& 4.050 \\

COT-TTS-0.6B-edit
& 0.6B
& 2.752
& 2.997
& 0.041
& 0.925
& 0.751
& 0.089
& 3.873
& 4.718
& 3.507
& 2.829
& 3.500 \\

COT-TTS-1.7B-edit
& 1.7B
& 2.763
& 3.127
& 0.054
& 0.953
& \textbf{0.557}
& 0.182
& 3.883
& 4.704
& 3.565
& 2.888
& 3.850 \\

\hline
\end{tabular}
}
\end{table*}

\begin{table*}[t]
\centering
\caption{Main results on the Chinese COT-TTS benchmark. ``$\uparrow$'' indicates that higher is better, and ``$\downarrow$'' indicates that lower is better.}
\label{tab:main_results_zh}
\resizebox{\textwidth}{!}{
\begin{tabular}{
l c
@{\hspace{8pt}}
c c c c c c
@{\hspace{12pt}}
c c c c
@{\hspace{12pt}}
c
}
\hline
\textbf{System}
& \textbf{Model Size}
& \multicolumn{6}{c}{\textbf{Speech Quality and Acoustic Consistency}}
& \multicolumn{4}{c}{\textbf{Context-Aware Reasoning Quality}}
& \multicolumn{1}{c}{\textbf{Human Evaluation}} \\
\cline{3-8}
\cline{9-12}
\cline{13-13}
&
& \shortstack{\textbf{UTMOSv2}\\$\uparrow$}
& \shortstack{\textbf{DNSMOSPro}\\$\uparrow$}
& \shortstack{\textbf{CER}\\$\downarrow$}
& \shortstack{\textbf{Emo. Cons.}\\$\uparrow$}
& \shortstack{\textbf{Dur. Err.}\\$\downarrow$}
& \shortstack{\textbf{F0 Corr.}\\$\uparrow$}
& \shortstack{\textbf{Hist. Und.}\\$\uparrow$}
& \shortstack{\textbf{CoT Logic}\\$\uparrow$}
& \shortstack{\textbf{Eff. Info. Richness}\\$\uparrow$}
& \shortstack{\textbf{LLM Composite}\\$\uparrow$}
& \shortstack{\textbf{Human MOS}\\$\uparrow$} \\
\hline

three-dia-a3b-fish
& 36.2B
& 2.872
& 3.303
& \textbf{0.010}
& 0.917
& 5.746
& 0.205
& \textbf{3.958}
& 4.585
& 3.096
& 2.402
& 3.150 \\

three-dia-a3b-voxcpm
& 34.2B
& 2.391
& 3.203
& 0.023
& 0.946
& 5.599
& 0.125
& 3.934
& 4.608
& 4.064
& 3.415
& 3.950 \\

three-qwen\_asr-a3b-fish
& 36.2B
& 2.874
& 3.279
& 0.055
& 0.921
& 5.286
& 0.188
& 3.708
& 4.789
& 2.932
& 2.234
& 3.700 \\

three-qwen\_asr-a3b-voxcpm
& 34.2B
& 2.372
& 3.155
& 0.056
& 0.943
& 5.248
& 0.140
& 3.747
& 4.798
& 3.942
& 3.282
& 3.200 \\

two-a3b-fish
& 39.0B
& 2.880
& 3.283
& 0.052
& 0.921
& 5.144
& \textbf{0.218}
& 3.531
& 4.812
& 2.545
& 1.847
& 3.150 \\

two-a3b-voxcpm
& 37.0B
& 2.378
& 3.156
& 0.058
& 0.946
& 5.225
& 0.123
& 3.546
& 4.803
& 3.822
& 3.101
& 3.350 \\

two-qwen3omni-seedvc
& $\sim$35.7B
& \textbf{2.974}
& \textbf{4.166}
& 0.063
& \textbf{0.955}
& 4.914
& 0.186
& --
& --
& --
& --
& 3.150 \\

\hline

COT-TTS-0.6B
& 0.6B
& 2.753
& 3.184
& 0.041
& 0.952
& 1.095
& 0.148
& 3.685
& 4.606
& 3.219
& 2.470
& \textbf{4.200} \\

COT-TTS-1.7B
& 1.7B
& 2.778
& 3.201
& 0.034
& \textbf{0.955}
& 0.944
& 0.202
& 3.671
& 4.698
& 3.156
& 2.425
& 4.150 \\

COT-TTS-0.6B-edit
& 0.6B
& 2.676
& 2.938
& 0.041
& 0.936
& 0.922
& 0.107
& 3.803
& \textbf{4.847}
& 4.079
& \textbf{3.461}
& 3.350 \\

COT-TTS-1.7B-edit
& 1.7B
& 2.656
& 3.113
& 0.034
& \textbf{0.955}
& \textbf{0.449}
& 0.142
& 3.845
& 4.712
& \textbf{4.092}
& 3.451
& 3.900 \\

\hline
\end{tabular}
}
\end{table*}

\subsection{Implementation and Inference Details}

Inference for both end-to-end models is performed on a single NVIDIA H800 80GB GPU with a batch size of 1. Each test sample is generated once using the same decoding configuration for the 0.6B and 1.7B models. If generation fails because of an execution error, one regeneration attempt is allowed. During inference, the model autoregressively generates the emotion-tagged historical transcript, CoT reasoning, and target speech tokens. The predicted semantic tokens are combined with the global tokens extracted from the reference speech for waveform reconstruction, as illustrated in Fig.~\ref{fig:e2e_model}. In the editable-CoT setting, an external LLM lightly revises the generated CoT while preserving its overall reasoning structure~\cite{deepseekai2025deepseekr1incentivizingreasoningcapability}. Detailed inference configurations are provided in the released code.

\section{Experimental Results}

\subsection{Comparison with Existing Systems} 

Tables~\ref{tab:main_results_en} and~\ref{tab:main_results_zh} present the main results on the English and Chinese COT-TTS benchmarks, respectively. The cascaded systems contain approximately 34--39B parameters in total, whereas our end-to-end models use only 0.6B or 1.7B parameters. Overall, our models achieve performance comparable to much larger cascaded baselines, while using substantially fewer parameters in a unified end-to-end framework. The detailed analysis of the evaluation results is presented below.

For objective speech metrics, the cascaded systems generally achieve higher UTMOSv2 and DNSMOSPro scores, especially with Fish Audio 2 and the AudioLLM--VC pipeline. Our end-to-end models show a clear advantage in duration consistency while maintaining competitive emotional-expression consistency and content preservation. Most cascaded baselines have duration errors above 5 seconds, while our models keep the error around 1 second or lower in most settings. Our models also achieve emotional-expression consistency scores around 0.95 and maintain low WER/CER. Under the editable-CoT setting, several speech-related metrics, including UTMOSv2, DNSMOSPro, and F0 correlation, decrease, although duration consistency is improved. This suggests that the model has learned a strong dependency between the generated CoT and subsequent speech realization, and large modifications to the CoT may move the reasoning context away from the distribution learned during training.

The LLM-based evaluation further shows that our end-to-end models remain competitive in several reasoning dimensions. Without CoT editing, both models already obtain competitive CoT-logic and historical-understanding scores. Editable CoT further improves the reasoning results, especially in historical understanding and effective information richness across both languages. On the Chinese benchmark, the edited models also achieve the highest LLM composite scores among all compared systems. These results show that our end-to-end models effectively understand historical audio context and perform context-aware speaking-manner reasoning. The gains from editable CoT are more pronounced on the Chinese benchmark, possibly because all CoT reasoning is generated in Chinese.

For human evaluation, listeners conduct a blind test in which they are provided with the historical dialogue audio, generated CoT, and synthesized target speech. The evaluation focuses on the appropriateness of the speaking manner with respect to the historical dialogue context, the consistency between the speech and the CoT reasoning, and the overall perceptual quality. The results show that the standard end-to-end models receive higher subjective scores than most cascaded baselines in both languages. After the blind evaluation, we conduct a post-hoc analysis with the system identities revealed. We observe that many baseline systems tend to maintain a similar speaking style throughout a long utterance, while our end-to-end models produce more natural variations in rhythm, emphasis, and emotion. The 1.7B model also tends to produce richer and stronger emotional variations than the 0.6B model. For TTS models controlled by style tags, a tag may affect the entire utterance even when the intended speaking manner applies only to a local segment. For example, a whisper instruction for a short phrase may cause the whole sentence to be synthesized in a whispering style. Qwen3-Omni can also occasionally deviate from the given target text and generate unrelated spoken content. However, the editable-CoT mode is sensitive to large modifications. Excessive modifications may disrupt the learned CoT--speech alignment and lead to artifacts such as trailing sounds, repetitions, or abnormally fast speech.

\begin{table*}[t]
\centering
\caption{Ablation results for the three-stage training strategy using the 0.6B model under non-edit inference. Each cell reports English / Chinese (EN / ZH) results. ``$\uparrow$'' indicates that higher is better, and ``$\downarrow$'' indicates that lower is better.}
\label{tab:training_ablation}
\resizebox{\textwidth}{!}{
\begin{tabular}{lccccccc}
\hline
\textbf{Training Setting}
& \textbf{UTMOSv2}
& \textbf{DNSMOSPro}
& \textbf{WER/CER}
& \textbf{Emo. Cons.}
& \textbf{Dur. Err.}
& \textbf{F0 Corr.}
& \textbf{Human MOS} \\
& $\uparrow$
& $\uparrow$
& $\downarrow$
& $\uparrow$
& $\downarrow$
& $\uparrow$
& $\uparrow$ \\
\multicolumn{8}{l}{\footnotesize Each cell is reported as EN / ZH. For the ASR error column, this corresponds to English WER / Chinese CER.} \\
\hline
Stage 1 only
& 2.539 / 2.492
& 2.734 / 3.064
& 0.549 / 0.467
& 0.907 / 0.928
& 6.323 / 6.163
& 0.086 / 0.105
& 2.150 / 2.250 \\

Stage 1 + Stage 2
& 2.619 / 2.544
& 2.900 / \textbf{3.222}
& 0.089 / 0.229
& 0.948 / 0.949
& 5.770 / 5.888
& 0.160 / 0.158
& 3.150 / 3.050 \\

Stage 1 + direct COT-TTS refinement
& 2.609 / 2.388
& 2.807 / 3.121
& 0.196 / 0.272
& 0.938 / 0.933
& 6.607 / 6.392
& 0.157 / 0.135
& 3.050 / 2.950 \\

Three-stage w/o auxiliary tasks
& 2.630 / 2.542
& 2.928 / 3.161
& 0.163 / 0.285
& 0.946 / 0.945
& 7.174 / 7.332
& 0.138 / 0.120
& 3.650 / 3.750 \\

Three-stage w/ quality-score conditioning
& 2.684 / 2.546
& 3.079 / 3.081
& 0.063 / 0.051
& 0.939 / 0.940
& 5.032 / 5.076
& \textbf{0.217} / \textbf{0.204}
& 3.600 / 3.550 \\

Full three-stage training (quality filtering)
& \textbf{2.943} / \textbf{2.753}
& \textbf{3.180} / 3.184
& \textbf{0.041} / \textbf{0.041}
& \textbf{0.954} / \textbf{0.952}
& \textbf{1.155} / \textbf{1.095}
& 0.168 / 0.148
& \textbf{4.150} / \textbf{4.200} \\
\hline
\end{tabular}
}
\end{table*}

\subsection{Ablation Study on Training Strategy}

We evaluate the three-stage training strategy using the 0.6B model under the standard non-edit inference setting. We compare the following training settings:

\begin{itemize}
    \item \textit{Stage 1 only}: only the modality-alignment stage is used.
    \item \textit{Stage 1 + Stage 2}: the final high-quality refinement stage is removed.
    \item \textit{Stage 1 + direct COT-TTS refinement}: Stage 2 is skipped, and the Stage-1 model is directly refined on the high-quality COT-TTS data.
    \item \textit{Three-stage w/o auxiliary tasks}: the three-stage pipeline is retained, but the auxiliary tasks are removed.
    \item \textit{Three-stage w/ quality-score conditioning}: UTMOSv2 and DNSMOSPro scores are used as additional conditions during Stage 3.
    \item \textit{Full three-stage training}: low-quality samples are removed before Stage-3 refinement, without using quality-score conditioning.
\end{itemize}

As shown in Table~\ref{tab:training_ablation}, Stage 1 alone is insufficient for COT-TTS. It provides basic speech understanding and synthesis ability, but the overall performance remains limited. Introducing Stage 2 leads to clear improvements in both objective evaluation and human MOS. Skipping Stage 2 and directly applying high-quality refinement gives weaker results. This shows that the large-scale COT-TTS training in Stage 2 cannot be replaced by the smaller high-quality subset. Removing the auxiliary tasks also degrades the overall performance, indicating that they provide useful complementary supervision during training. The largest gains appear after Stage 2, suggesting that large-scale task-specific training is important for establishing the core COT-TTS capability before high-quality refinement.

The design of Stage 3 further affects the final model. Using UTMOSv2 and DNSMOSPro scores as control conditions provides only limited improvement. In contrast, directly removing low-quality samples gives better overall results and achieves the highest human MOS in both languages. The quality-filtering strategy also improves recognition accuracy and duration consistency, showing benefits beyond perceptual quality alone. This suggests that improving the training data itself is more effective than introducing estimated quality scores as additional conditions. Therefore, we adopt quality filtering as the final Stage-3 training strategy.

\subsection{Qualitative Analysis and Web Demo}

We first analyze two representative cases with complete model inputs and outputs, both of which are available on our demo page. In the new-speaker response case, the father blames himself for causing trouble for his daughter. The model reasons that the daughter should first reassure her father and then explain the current situation. The reference speech carries a substantially different emotion from the generated speech. However, the speaker timbre remains similar. This suggests that the reference speech mainly provides speaker conditioning, while the emotional expression is determined by the dialogue context and CoT reasoning. In the same-speaker continuation case, the speaker discusses why the chancellor should not yet claim the throne, expressing dissatisfaction and helplessness. The model maintains the preceding emotion and speaking rhythm, while expressing concern about losing public support. These cases show that the model can distinguish between response and continuation and infer appropriate emotions from the context.

\begin{table}[!htbp]
\centering
\caption{Editable-CoT settings for controlled speech synthesis. Only the specified CoT fields are modified.}
\label{tab:editable_cot}
\scriptsize
\setlength{\tabcolsep}{2pt}
\begin{tabularx}{\columnwidth}{l*{5}{>{\centering\arraybackslash}X}}
\hline
\textbf{Control} & \textbf{Default} &
\multicolumn{4}{c}{\textbf{Edited Parameters}} \\
\hline
Duration (s)
& 5.54 & 4.5 & 6.5 & 7.5 & 8.5 \\

Rhythm (s)
& 4.23/5.54 & 3.2/4.1 & 3.7/4.8 & 4.9/6.4 & 5.6/7.5 \\

Intensity
& 0.77 & 0.57 & 0.67 & 0.87 & 0.97 \\
\hline
\end{tabularx}
\end{table}

We next examine the controllability of editable CoT. We keep the historical audio, target text, and reference speech fixed, and edit only selected CoT fields. As shown in Table~\ref{tab:editable_cot}, we vary the total duration, rhythm, and emotional expression intensity. For rhythm control, we adjust the valid speech duration and total duration. The generated speech changes clearly and consistently with these edits, demonstrating effective control over the selected speaking attributes. The corresponding audio examples are available on our demo page. We also observe several practical conditions for effective CoT editing. Moderate edits generally produce more stable results, while large changes may introduce noise, repetition, or word omission. A very low valid-speech ratio can also increase the risk of missing content. The edited dimensions should remain internally consistent. For example, describing the emotion as slightly angry in one dimension but extremely angry in another can weaken the control effect. Different attributes may also interact. Stronger anger, for example, is often associated with a shorter duration. Therefore, CoT editing should maintain consistency across dimensions and preserve coherent internal reasoning. Finally, the edited CoT should remain compatible with the dialogue context. Imposing a happy emotion in an angry scene is therefore difficult even when the corresponding CoT fields are modified.

We further examine emotional expressiveness under different dialogue contexts. We select five emotion categories: anger, sadness, steady confidence, proud confidence, and surprise. Each category contains one Chinese and one English example. The generated speech shows clear and appropriate emotional characteristics across these cases. The model also produces natural variations in emotion, emphasis, and speaking rhythm according to the dialogue context. More qualitative examples are available on our demo page at \url{https://luckybian.github.io/COT-TTS}.

\section{Conclusion}

In this paper, we introduced \emph{COT-TTS}, a context-aware reasoning TTS task that infers how a fixed target utterance should be spoken from historical dialogue audio. We developed a reproducible data construction pipeline, yielding 9M bilingual training samples with a 1M high-quality subset, and built a source-disjoint benchmark of 800 human-validated samples. We also established an evaluation protocol covering both speech synthesis and context-aware reasoning. We further developed 0.6B- and 1.7B-parameter end-to-end autoregressive models that generate speaking-manner reasoning before target speech synthesis. The intermediate CoT can be inspected and edited during inference. Our qualitative analysis demonstrates control over duration, rhythm, and emotional expression through CoT editing. The models also produce expressive speech across different dialogue contexts. Experimental results show that our models achieve performance comparable to much larger cascaded systems with substantially fewer parameters. Our models also achieve higher human subjective evaluation scores than most cascaded baselines. The proposed benchmark provides a unified setting for evaluating context-aware reasoning and speech synthesis. The explicit CoT also offers a more interpretable interface for analyzing speech realization. Future work will focus on improving editable-CoT robustness and finer-grained control over speech realization.

\bibliographystyle{IEEEtran}
\bibliography{references}

\end{document}